%% file: main.tex
\documentclass{article}
\usepackage[a4paper, total={6in, 8in}]{geometry}

\usepackage{algorithm}
\usepackage{algpseudocode}
\usepackage{amsmath,nicefrac}
\usepackage{graphicx,amssymb,amsthm,xcolor,booktabs,url,hyperref}
\usepackage{authblk}

\newtheorem{proposition}{Proposition}
\newtheorem{definition}{Definition}

\newcommand{\Description}[1]{}
\input{main/notations}

\title{Improving Unlearning with Model Updates\\ Probably Aligned with Gradients}

\author[1]{Virgile Dine\footnote{contact author: \texttt{virgile.dine@inria.fr}\\ paper accepted to AISec'25 co-located with the 32nd ACM Conference on Computer and Communications Security}}
\author[1]{Teddy Furon}
\author[2]{Charly Faure}
\affil[1]{Centre Inria de l'Université de Rennes, France}
\affil[2]{AMIAD, France}
\date{}

\begin{document}
\maketitle

\begin{abstract}
    \input{main/0_abstract}
\end{abstract}

\section{Introduction}
\label{sec:introduction}
\input{main/1_introduction}

\section{Related Work}
\label{sec:relatedwork}
\input{main/2_related_work}

\section{Problem Statement}
\label{sec:problemstate}
\input{main/3_problem_state}

\section{General Framework for First-Order Methods}
\label{sec:kkt}
\input{main/4_modelization}

\section{Processing with Batches}
\label{sec:minibatchs}
\input{main/5_statistical}

\section{Experiments}
\label{sec:experiments}
\input{main/tables/table_sub_1}

\input{main/6_experiments}
\input{main/6_results}

\section{Discussion and Limitations}
\input{main/7_discussion}

\clearpage

\bibliographystyle{plain}
\bibliography{main/bib}

\appendix
\section*{Appendix: Proofs of the Propositions}
\label{appendix:proofs}
\input{main/appendix_proof}

\end{document}

%% file: main/notations.tex
\def\Exp{\mathbb{E}}
\def\Prob{\mathbb{P}}
\def\IdealUModel{\theta^\star}
\def\sD{\mathcal{D}}

\def\sX{\mathcal{X}}
\def\sY{\mathcal{Y}}

\def\TRAIN{\mathrm{train}}
\def\TEST{\mathrm{test}}
\def\FORGET{\mathrm{forget}}
\def\RETAIN{\mathrm{retain}}

\def\loss{L}
\def\con{C}

\def\sif{\mathrm{Agg}}
\def\ber{\mathcal{B}}

\def\ie{\textit{i.e.}}

\def\AND{\mathrm{AND}}
\def\PROB{\mathrm{PROB}}
\def\BER{\mathrm{BER}}
\def\FOCUS{\mathrm{F}}

\def\focus{f}

\def\siz{0.75}
\def\rat{0.75}

%% file: main/0_abstract.tex

We formulate the machine unlearning problem as a general constrained optimization problem.
It unifies the first-order methods from the approximate machine unlearning literature.
This paper then introduces the concept of feasible updates as the model's parameter update directions that help with unlearning while not degrading the utility of the initial model. Our design of feasible updates is based on masking, \ie\ a careful selection of the model's parameters worth updating.
It also takes into account the estimation noise of the gradients when processing each batch of data to offer a statistical guarantee to derive locally feasible updates.
The technique can be plugged in, as an add-on, to any first-order approximate unlearning methods.
Experiments with computer vision classifiers validates this approach.

%% file: main/1_introduction.tex

Machine learning models are integrated into many real-world applications.
Since these models contain artifacts of potentially sensitive training data, this raises concerns about data confidentiality and user privacy.
The ability to remove specific training data from a model has emerged as a key mechanism to enforce, for instance, the "right to be forgotten" promoted by the European GDPR law~\cite{Hoofnagle2019TheEU} or the "right to erase" in the Canadian CPPA legislation~\cite{CanadaLaw}.
Approximate machine unlearning aims to find efficient mechanisms, avoiding the cost of learning a new model from scratch over the training dataset deprived of the sensitive data.

Privacy is not the only application of machine unlearning.
It has been proven useful as a defense against backdoor attacks by annihilating the influence of the poisoned training data~\cite{zhang2024backdoor,Wang2019NeuralCI}, or to improve fairness by removing data that induce biases in the training set.
Another scenario is the derivation of a restricted public model from a powerful private model learned on some sensitive data~\cite{golatkar2020eternal}.  
The accuracy of the public model should be on par with or slightly degraded compared to the private model.
On the other hand, the model should not leak information about the sensitive training data. The procedure used by an attacker to infer whether a specific data record was included in the training set is known as a Membership Inference Attack (MIA)~\cite{shokri2017membership}.

This mechanism inherently provokes tension between the goal of forgetting some training data and the non-desired consequence of unlearning the model's capabilities.
Indeed, the term \emph{Machine Unlearning} is somewhat misleading: it unlearns some training data, meaning forgets, not its ability to perform a given task (coined as \emph{Catastrophic Forgetting} in machine unlearning).

This work assumes that the unlearner is also the creator of the initial classifier.
It means that the unlearner knows the initial model and all its parameters, the dataset to retain, and the dataset to forget.
The unlearner aims to find an update of the initial model's parameters that erases the influence of the forget dataset.

Our first contribution formalizes machine unlearning as a constrained optimization problem: forgetting the data while sustaining the model's utility.
This generalizes first-order unlearning methods through a single framework.
A second contribution introduces the concept of feasible update as a sound elementary solution to the constrained optimization problem.  
We propose a general procedure to craft feasible updates based on masks, carefully selecting the parameters to be updated.
These are presented as add-ons to be plugged into the existing unlearning methods covered by our framework.
A third contribution takes into account the processing per batch of data.
A simple statistical model relating the gradient computed over a batch to the gradient computed over the whole dataset gives birth to a statistical guarantee that the mask selects the appropriate parameters. The appendix contains the proofs of our propositions. To summarize, this work provides a theoretical foundation for masking unlearning methods like saliency unlearning~\cite{Fan2023SalUnEM}.  
Our approach is validated over 360 experimental configurations combining several add-ons, unlearning methods, classifiers, datasets, types of unlearning scenarios, and sizes of the portion of data that must be unlearned.

%% file: main/2_related_work.tex

\subsection{Machine Unlearning}
\label{sec:RelWorkUnlearn}

This work deals with centralized machine unlearning, which starts with a regular neural network classification model and obtains a model sharing the same architecture but with updated parameters. It excludes techniques that prepare the unlearning actually before the learning by resorting to specific data structures and model architectures, like split unlearning techniques~\cite{golatkar2021mixed, yan2022arcane, graves2021amnesiac, garg2020formalizing, chen2022recommendation,bourtoule2021machine}.  
These techniques indeed target exact unlearning, which makes the output of the unlearned model statistically indistinguishable from the output of the model that never saw the forget data during its training~\cite{xu2023machineunlearningsurvey}.

On the contrary, our proposal pertains to the approximate unlearning strategy and, more specifically, to fast first-order methods that compute one or more gradients to proceed a model's parameters update. This excludes the more demanding second-order methods that compute Hessian matrices~\cite{guo2020certified}.

Among the first-order method family, one can find the following basic schemes: Fine-Tuning (FT)~\cite{Warnecke2021MachineUO} where the model is further refined using only the remaining dataset $\sD_\RETAIN$, Gradient Ascent (GA)~\cite{Thudi2021UnrollingSU} which attempts to undo learning by applying gradient ascent on the forget dataset $\sD_\FORGET$, and Successive Random Labeling (SRL)~\cite{golatkar2020eternal} which fine-tunes the model over the whole dataset $\sD_{\TRAIN}$ but with random labels for data belonging to $\sD_\FORGET$.
Slightly more elaborated are bi-objective schemes such as Negative Gradient Plus (NGPlus)~\cite{kurmanji2023towards}, a mixed version of gradient ascent and fine-tuning that performs a gradient ascent over $\sD_\FORGET$ with a descent over $\sD_\RETAIN$. It also encompasses methods considering the gradient of functions that are not classification losses. For example sparse-MU~\cite{jia2023model}, which integrates the $\ell_1$-norm of the model parameter as a penalty. SCalable Remembering and Unlearning unBound (SCRUB)~\cite{kurmanji2023towards}, minimizes (resp. maximizes) the Kullback-Leibler divergence between the predicted probabilities of the initial and the unlearned models over $\sD_\RETAIN$ (resp. $\sD_\FORGET$).

Finally, our work is inspired by the Saliency Unlearning method (SalUn)~\cite{Fan2023SalUnEM}, which enforces the sparsity of the parameter update by a binary mask.
It amounts to selecting some parameters to be updated and leaving the remaining ones untouched.
The selection is driven by the amplitude of the components of the gradient of the classification loss over $\sD_\FORGET$.
Our work proposes a theoretical framework for designing more explainable masks.

Interested readers may find many more references in the following surveys~\cite{Shaik2024exploring,xu2023machineunlearningsurvey,wang2024machine}.

\subsection{Membership Inference Attack}
\label{sec:MIA}
A Membership Inference Attack (MIA) discloses whether a given piece of data belongs to the training set of a model.
In other words, whether the model saw this piece of data during its training.
The evaluation of an unlearning technique uses an MIA to verify that the data of $\sD_\FORGET$ are no longer deemed as training data.

MIAs are usually intensive because they compute shadow models, \ie\ a set of models trained with and without this piece of data in order to model how the membership impacts the outputs of a classifier like~\cite{shokri2017membership} or the LiRa attack~\cite{carlini2022membership}.  
U-LiRa~\cite{hayes2024inexact} is an attack more adapted to machine unlearning that distinguishes data unlearned or not in the training set.
Lightweight alternative MIAs (dedicated to unlearning or not) refuse the cost of the shadow models' computation.
They take decisions based on simple statistics from the model output~\cite{hu2022membership, yeom2018privacy, li2022leaks, bertran2023scalable, choquette2021label,jia2023model,kurmanji2023towards}.

Even though U-LiRa is more effective, the experiments reported in~\cite{hayes2024inexact} reflect that the MIAs lead to similar evaluations: when U-LiRa measures that one unlearning method is better than another, another MIA generally leads to the same conclusion.
Our paper proposes versatile add-ons that can be plugged into many unlearning techniques. 
Our objective is to show the impact due to our add-ons.
Therefore, we can resort to simple MIAs to measure a relative improvement w.r.t. the baseline technique. 

\subsection{Gradient Agreements}
\label{sec:MTL}
This article frames unlearning as a constrained optimization problem.
Similarly to Multi-Task Learning with two tasks, the issue is to combine two gradients (w.r.t. the model parameters) into a single parameter update vector.
Some well-known aggregations are IGA~\cite{koyama2020out}, or PCGrad~\cite{yu2020gradient} and~CAGrad \cite{liu2021conflict} that modify gradients by projecting one onto another.
Some recent proposals induce a sparse aggregation.
The $\AND$ masking strategy~\cite{andmask} verifies whether all gradients point in the same direction and updates a network parameter only if there is full agreement among them.
This masking approach is similar to using a logical $\AND$ operation in gradient directions. 
The idea was improved~\cite{sandmask} to tackle real-world problems where the collected data differs with respect to some environment.
Our work rediscovers the $\AND$ formulation but from a different rationale stemming from the KKT conditions (see Prop.~\ref{prop:equilibrium}).

%% file: main/3_problem_state.tex
\subsection{Notations}
\label{sec:Notations}
Datasets are noted with a calligraphic font.
For a given function $F$, $\Exp_{s\sim\sD}[F(s)]$ is the expectation of the random variable $F(s)$ when $s$ follows a uniform distribution over the finite set $\sD$.
In the same manner, $\Prob_{s\sim\sD}[F(s)=y]$ is the probability that the output of $F$ equals a given $y$. 

We denote by $u \odot v \doteq (u_i v_i)_i$ the vector corresponding to the term-by-term product of two vectors $u = (u_i)_i$ and $v=(v_i)_i$.
We write $u\leq 0$ (resp. $\geq 0$) when the vector $u$ verifies $\forall i,\, u_i \leq 0$ (resp. $\geq 0$).
This enables us to use inequalities directly on vectors when it makes sense.

For the sake of simplicity, we note $F\odot G \doteq s\mapsto F(s)\odot G(s)$ and equations are written function-wise whenever possible, \ie\ $F=G$ when $\forall s,\, F(s) = G(s)$, or $M$ for a model $M_\theta$ if there is no ambiguity.

\subsection{Exact and Approximate Unlearning}
\label{sec:ExactApprox}
Consider a trained classifier $M_{\theta_0}: \sX \to \sY$ predicting labels in the set $\mathcal{Y}$ from observations in the input space $\mathcal{X}$. 
Supervised learning found the model's parameters $\theta_0$ by optimizing a loss function $L$ over a training set $\sD_{\TRAIN} \subset \mathcal{X}\times\mathcal{Y}$:
\begin{equation}
        \theta_0 \in \arg \min_{\theta\in\Theta} \Exp_{(x,y)\sim \sD_{\TRAIN}} \left[L (M_\theta(x), y)\right].
        \label{eq:InitialModel}
\end{equation}
Data forgetting consists in unlearning some portion $\sD_\FORGET \subset \sD_{\TRAIN}$ of the data while conserving performance as if the model had been trained over the set to retain $\sD_\RETAIN = \sD_\TRAIN \setminus \sD_\FORGET$.
Then \textit{exact unlearning} amounts to finding the parameters $\IdealUModel$ of a new model as if it was trained from scratch without the problematic data $\sD_\FORGET$ as follows:

\begin{equation}
        \IdealUModel \in \arg\min_{\theta\in\Theta} \Exp_{(x,y)\sim \sD_\RETAIN} \left[L (M_\theta(x), y)\right].
        \label{unlearningprob}
\end{equation}
\textit{Approximate unlearning} methods consist of low-cost modification of the trained parameters $\theta_0$ so that the unlearned model is close to the exact one~\eqref{unlearningprob}, which has never seen the set $\sD_\FORGET$.

\subsection{Usual Evaluation Metrics}
\label{sec:EvalMetrics}
In real-world unlearning scenarios, the parameters $\IdealUModel$ of the exact model is unknown, preventing any measure of closeness in the parameter space $\Theta$. Indeed, it remains to be defined how the unlearned model should mimic the behavior of the exact one, $M_{\theta^{\star}}$.
This is usually measured by common accuracy metrics ~\cite{jia2023model} such as:
\begin{equation}
        \mathbb{P}_{(x,y)\sim \sD}\left[ M(x) = y \right],
        \label{eq:AccDef}
\end{equation}
with $\sD = \sD_\FORGET$ for the Unlearning Accuracy (UA); $\sD = \sD_\RETAIN$ for the Retain Accuracy (RA); and $\sD = \sD_\TEST$, a test set with empty intersection with $\sD_\TRAIN$, for the Test Accuracy (TA).

Another important criterion is the attack success rate of MIAs. Verifying if the binary decision rule $b$, trained to output $1$ if the piece of data $x$ is deemed as belonging to the training set and $0$ otherwise, has a good accuracy over the forget set:
\begin{equation}
        \mathrm{MIA} \doteq \mathbb{E}_{(x,y)\sim \sD_\FORGET}\left[ b(M,  x) \right].
        \label{eq:MIADef}
\end{equation}
The lower, the better. If this score is low, the decision rule considers the forgotten data as test data. If it is high, the decision rule considers the forgotten data as part of the training data. A value near $\nicefrac{1}{2}$ means that the decision rule cannot assign the whole forgotten data to be part of the training or test data.

Last but not least, the Run Time Efficiency (RTE) measures the duration of the unlearning procedure.

Our experimental protocol introduces two other metrics (see Sect.~\ref{sec:EvaluationMetrics}) that are rarely used in the literature because they require the computation of the ideal model $M_{\IdealUModel}$~\eqref{unlearningprob}.

%% file: main/4_modelization.tex
This section proposes a framework that encompasses the bi-objective first-order unlearning methods proposed in the literature and that leads to the concept of feasible update.

\subsection{Optimization Problem}
\label{sec:GenFrame}
Let $M_{\theta_0}$ be the initial model~\eqref{eq:InitialModel} trained to minimize the loss $L$ over the dataset $\sD_\TRAIN$, and $\sD_\FORGET\subset\sD_\TRAIN$ the dataset to be forgotten.
We model the unlearning as a constrained optimization problem:

\begin{equation}
    \begin{split}
    & \min_{\theta\in\Theta} U(\theta)\\
        \textit{s. t.}\quad & C(\theta) \leq 0,
    \end{split}
    \label{problem}
\end{equation}
where the functional $U$ drives the unlearning objective while $C$ encodes the preservation of the classification performance. These functions are not always, but typically, defined as:

\begin{eqnarray}
 U(\theta)&\doteq&  \Exp_{(x,y)\sim \sD_U}\left[\loss_U\left(M_\theta,x, y\right)\right],\\
 C(\theta)&\doteq&  \Exp_{(x,y)\sim \sD_C}\left[\loss_C(M_\theta,x, y)-\loss_C(M_{\theta_0},x, y)\right],
 \label{eq:Constraint}
\end{eqnarray}
where $\sD_U$ and $\sD_C$ are two datasets, and $\loss_U$ and $\loss_C$ are two losses.
The constraint means that the parameter $\theta$ shall not deteriorate the performances of the initial model, at least over the subset $\sD_C$.
The formulation~\eqref{problem} is general as it encompasses the first-order methods introduced in Sect.~\ref{sec:RelWorkUnlearn}:

\begin{itemize}
    \item FT (Fine-tuning): set $\loss_U = L$ with $\sD_U=\sD_\RETAIN$, and $\con$ the null function (no constraint),
    
    \item GA (Gradient Ascent): set $\loss_U = -L$ with $\sD_U=\sD_\FORGET$, and $\con$ the null function (no constraint),
    
    \item NGPlus (Negative Gradient Plus): set $\loss_U = -L$ with $\sD_U = \sD_\FORGET$, and $\loss_C = L$ with $\sD_C= \sD_\RETAIN$,
    
    \item SRL (Successive Random Labeling): set $\loss_U = L$ over a special dataset $\sD_U$ where $y$ is replaced by a random label when $(x,y)\in \sD_\FORGET$, and $\loss_C = L$ with $\sD_C=\sD_\RETAIN$,
    
    \item $\ell_1$-sparse MU: set $\loss_U = \|\theta\|_1$ with $\sD_U = \sD_\RETAIN$, and $\loss_C = L$ with $\sD_{C} = \sD_\RETAIN$,
    
    \item SCRUB: set $\loss_U = -\mathrm{KL}(M_{\theta_0}\| M_\theta)$ with $\sD_U = \sD_\FORGET$, and $\loss_C = \mathrm{KL}(M_{\theta_0}\| M_\theta) + \gamma \cdot L$ with $\sD_C= \sD_\RETAIN$.
\end{itemize}

\subsection{KKT Condition at Equilibrium}
\label{sec:KKT}
A classical result in constrained optimization allows us to relate the gradients of the unlearning objective and constraint functions.

\begin{proposition}[Necessary Condition at Equilibrium]
    \label{prop:equilibrium}
    In the optimization problem \eqref{problem}, the KKT conditions impose the necessary condition at the equilibrium $\bar{\theta}$:
    \begin{equation}
        \nabla U \odot \nabla C \left(\bar{\theta}\right) \leq 0.
    \end{equation}
\end{proposition}

The Prop.~\ref{prop:equilibrium} is easily interpreted: at a local minimum $\bar\theta$ there is no longer a direction in the parameter space to improve $U$ while maintaining performance $C$.

\subsection{Feasible Update}
\label{sec:convenient}
This section first gives a desirable property of an update of the model parameter $\theta\leftarrow\theta+\eta\Delta$, and then proposes a design.

\begin{definition}
    The update direction $\Delta \in \Theta$ is said to be
    \emph{feasible} if it is negatively aligned with both gradients $\nabla U(\theta)$ and $\nabla C(\theta)$:
    \begin{equation}
        \label{eq-convenient}
        \langle \Delta,\nabla U(\theta) \rangle \leq 0 \quad\text{and}\quad
        \langle \Delta,\nabla C(\theta) \rangle \leq 0.
    \end{equation}
\end{definition}

The goal is simple: such an update, up to the first order, minimizes the loss function $U$ while decreasing the constraint $C$. 

\begin{proposition}[Feasible update's Guarantee]
    \label{prop:guarantee}
    Any feasible update~\eqref{eq-convenient} ensures to fulfill the constraint while improving the minimization of the problem~\eqref{problem}.
    That is, for any feasible $\Delta$, there exists a sufficiently small learning rate $\eta > 0$ so that starting at $\theta_0$ one has
    \begin{equation}
        U(\theta_0+\eta\Delta) \leq U(\theta_0) \text{ and }
        C(\theta_0+\eta\Delta) \leq 0.
    \end{equation}
\end{proposition}

Our design for a feasible update is based on two parts: a special weighting of gradients and an aggregation $\sif$.
Following the necessary condition \ref{prop:equilibrium}, we define a selection of the parameters to be updated. This takes the form of a boolean mask:

\begin{equation}
    m_{\AND}(\nabla U,\nabla C)_i \doteq
    \begin{cases}
        1 & \text{if}\,\, \nicefrac{\partial}{\partial\theta_i} U \cdot \nicefrac{\partial}{\partial\theta_i} C > 0, \\
        0 & \text{otherwise}.
    \end{cases}
    \label{eq:Mask}
\end{equation}
The interpretation is also simple: the mask indicates the parameters that we can safely update.
Up to the first order, a change of one of these parameters modifies both $U$ and $C$ in the same way, whether it is an increase or a decrease.

From the construction of the mask, it is easy to design a feasible update by combining both gradients.
We define a sign-invariant function $\sif:\mathbb{R}\times\mathbb{R}\to\mathbb{R}$.
The following construction is feasible as it ensures a negative correlation with the gradients:
\begin{equation}
\label{eq:Aggreg}
    \Delta_{\AND} \doteq - m_{\AND}\odot \sif\left(\nabla U,\nabla C\right),
\end{equation}
with an abuse of notation where the equation is function-wise (\ie, $\theta$ is omitted for the sake of simplicity) and where function $\sif$ is applied component-wise.
As for the choice of this sign-invariant function, we propose:
\begin{align}
    \sif(x,y) &= \alpha x + \beta y, \quad(\alpha,\beta)\in\mathbb{R}^2_{+}, 
    \label{eq:linG}\\
    \sif(x,y) &= \underset{z \in \{x, y\}}{\operatorname{argmin}} \,|z|.
    \label{eq:MinG}
\end{align}
The last suggestion is indeed the aggregation choice made in the Mask-Small-Gradients (MSG) method~\cite{msg} to compute the random re-initialization before the fine-tuning phase.

\begin{figure}[t]
    \centering
    \includegraphics[width=\rat\columnwidth]{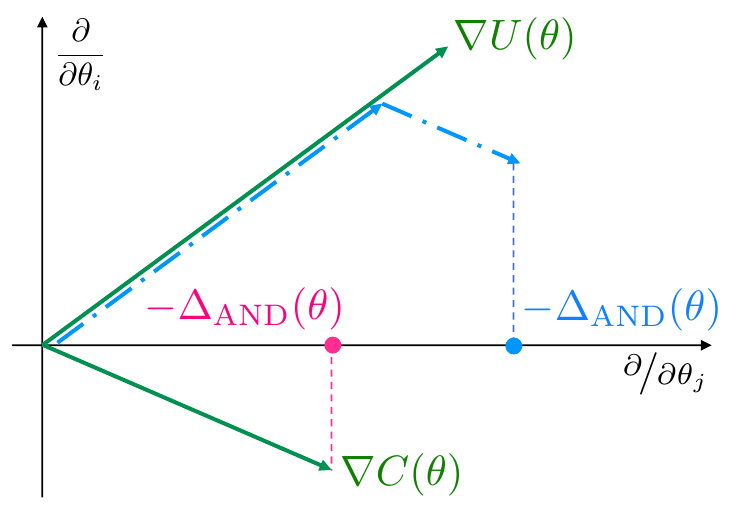}
    \caption{Masking procedure to ensure agreement between losses with $\sif$~\eqref{eq:MinG} in pink and~\eqref{eq:linG} in blue ($\alpha=0.8$, $\beta=0.5$).} 
    \label{fig:masking}
    \Description{Masking procedure to ensure agreement between losses with $\sif$~\eqref{eq:MinG} in pink and~\eqref{eq:linG} in blue ($\alpha=0.8$, $\beta=0.5$).}
\end{figure}

Our technique belongs to the first-order methods family that makes a small parameter shift. Additionally, its amplitude can be quantified.

\begin{proposition}[Vicinity of $\theta_0$]
    \label{prop:bounding}
    Denoting the number of parameters selected by any mask $m$ by $\|m(\nabla U,\nabla C)\|_0$ and the update direction $\Delta\doteq - m\odot\sif(\nabla U,\nabla C)$. Then the updated parameters $\theta = \theta_0 + \eta\Delta(\theta_0)$, is close to the original parameter: $\forall q\geq 1$
    \begin{equation}
        \|\theta - \theta_0 \|_q \leq \eta\|m(\nabla U,\nabla C)\left(\theta_0\right)\|_0^{\nicefrac{1}{q}}\|\Delta(\theta_0) \|_\infty.
    \end{equation}
    Moreover, if the update is based on the absolute min function~\eqref{eq:MinG}:
    \begin{equation}
        \label{boundedtheta}
        \|\theta - \theta_0 \|_q \leq \eta\|m(\nabla U,\nabla C)\left(\theta_0\right)\|_0^{\nicefrac{1}{q}}\| \nabla C(\theta_0)\|_\infty.
    \end{equation}
\end{proposition}

It turns out that if $\sD_C \subset \sD_\TRAIN$ then $\|\nabla C(\theta_0)\|_{\infty} \simeq 0$ because the constraint is usually the classification loss over the data to be retained, \ie\ data already seen during the training of $M_{\theta_0}$. Besides, experimentally $\|m_{\AND}(\nabla U,\nabla C)\left(\theta_0\right)\|_0$ happens to decrease so that the update is at each step a slighter change of the parameters. As models are Lipschitz w.r.t. those parameters~\cite{virmaux2018lipschitz, fazlyab2019efficient}, the output of the updated model is never drastically different than the output of the initial model.

%% file: main/5_statistical.tex
The previous section explains an add-on technique, assuming that the unlearning procedure has access to functions $U$ and $C$ and their gradients.
This section keeps the same optimization problem~\eqref{problem} but without access to the true gradients $\nabla U$ and $\nabla C$ and we use new notations to clarify this distinction.
This illustrates the case where $U$ and $C$ are averages of classification losses over large datasets. Machine learning practitioners typically address this issue by computing gradients over batches. 

\subsection{Probabilistic Modeling of Gradients}
\label{sec:statmodel}
We consider a classic statistical model: computed over a random batch, the gradient $\hat g_U(\theta)$ (resp. $\hat g_C(\theta)$) is a noisy estimation of the true gradient $g_U(\theta) \doteq \nabla U(\theta)$ (resp. $g_C(\theta) \doteq \nabla C(\theta)$):
\begin{equation}
    \label{eq:StatModel}
    \begin{split}
        \hat g_U &= g_U + N_U, \text{with}\quad N_U\sim\mathcal{N}(0;\Sigma_U),\\
        \hat g_C &= g_C + N_C, \text{with}\quad N_C\sim\mathcal{N}(0;\Sigma_C),
    \end{split}
\end{equation}
where $N_U$ and $N_C$ are two independent Gaussian random vectors.

\begin{figure}[tbh]
    \centering
    \includegraphics[width=\rat\columnwidth]{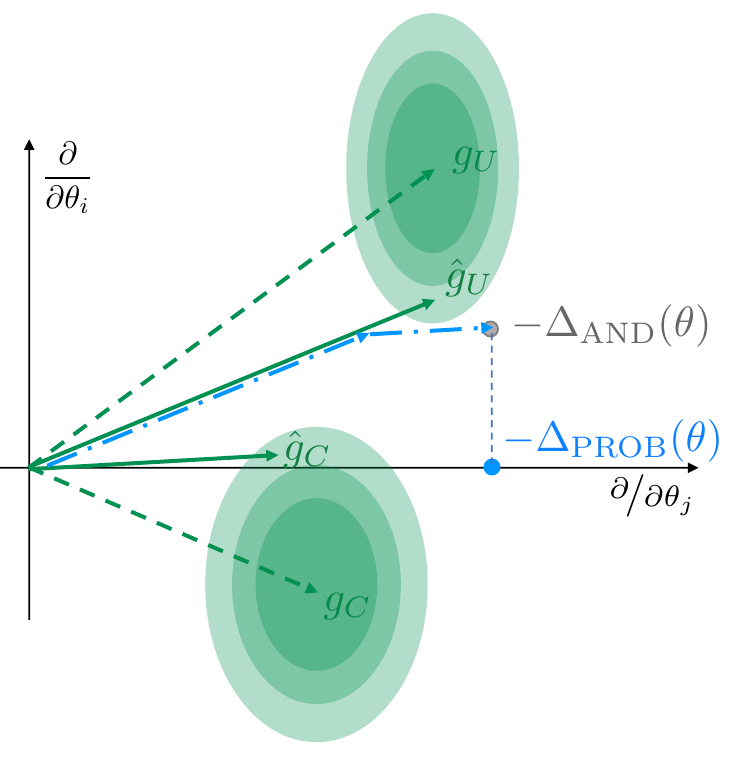}
    \caption{Masking wrong modification direction caused by noisy gradient, with $\sif$~\eqref{eq:linG} for the updates ($\alpha=0.8$, $\beta=0.5$).}
    \Description{Masking wrong modification direction caused by noisy gradient, with $\sif$~\eqref{eq:linG} for updates ($\alpha=0.8$, $\beta=0.5$).}
    \label{fig:maskingSalUn}
\end{figure}

The estimation noise may induce a modification of some parameters that the true gradients do not recommend (see Fig.~\ref{fig:maskingSalUn}). Our conservative strategy is to make the selection process by the mask~\eqref{eq:Mask} as certain as possible. This amounts to evaluating the probability of a correct selection of the $i$-th parameter, based on the noisy components $\hat g_{U,i}$ and $\hat g_{C,i}$ of the gradients observed at a given batch. 

\begin{proposition}[Gradient's Probable Alignment]
    Given the statistical model~\eqref{eq:StatModel} and a batch over which the empirical gradients take the value $\hat{g}_U$ and $\hat{g}_C$, the selection of the $i$-th parameter is a good choice with probability:
    \begin{equation}
        \Prob\left[g_{U,i}g_{C,i} > 0 \,\big|\, \hat g_{U}, \hat g_{C}\right] = \phi_{U,i}\cdot\phi_{C,i}+(1-\phi_{U,i})(1-\phi_{C,i}),
        \label{eq:AgreeProb}
    \end{equation}
    where the vector $\phi_U \doteq \Phi(\nicefrac{\hat g_U}{\sqrt{\mathrm{diag}(\Sigma_U)}})$ (idem for $\phi_C$) and the operators are taken component-wise. $\Phi$ is the cumulative distribution function of a standard Gaussian: $\Phi(x) \doteq \frac{1}{\sqrt{2\pi}}\int_{-\infty}^x {\exp(-\frac{t^2}{2})}\,\mathrm{d}t$.
    \label{prop:diragreeprob}
\end{proposition}

\begin{figure}[!b]
    \centering
    \includegraphics[width=\rat\columnwidth]{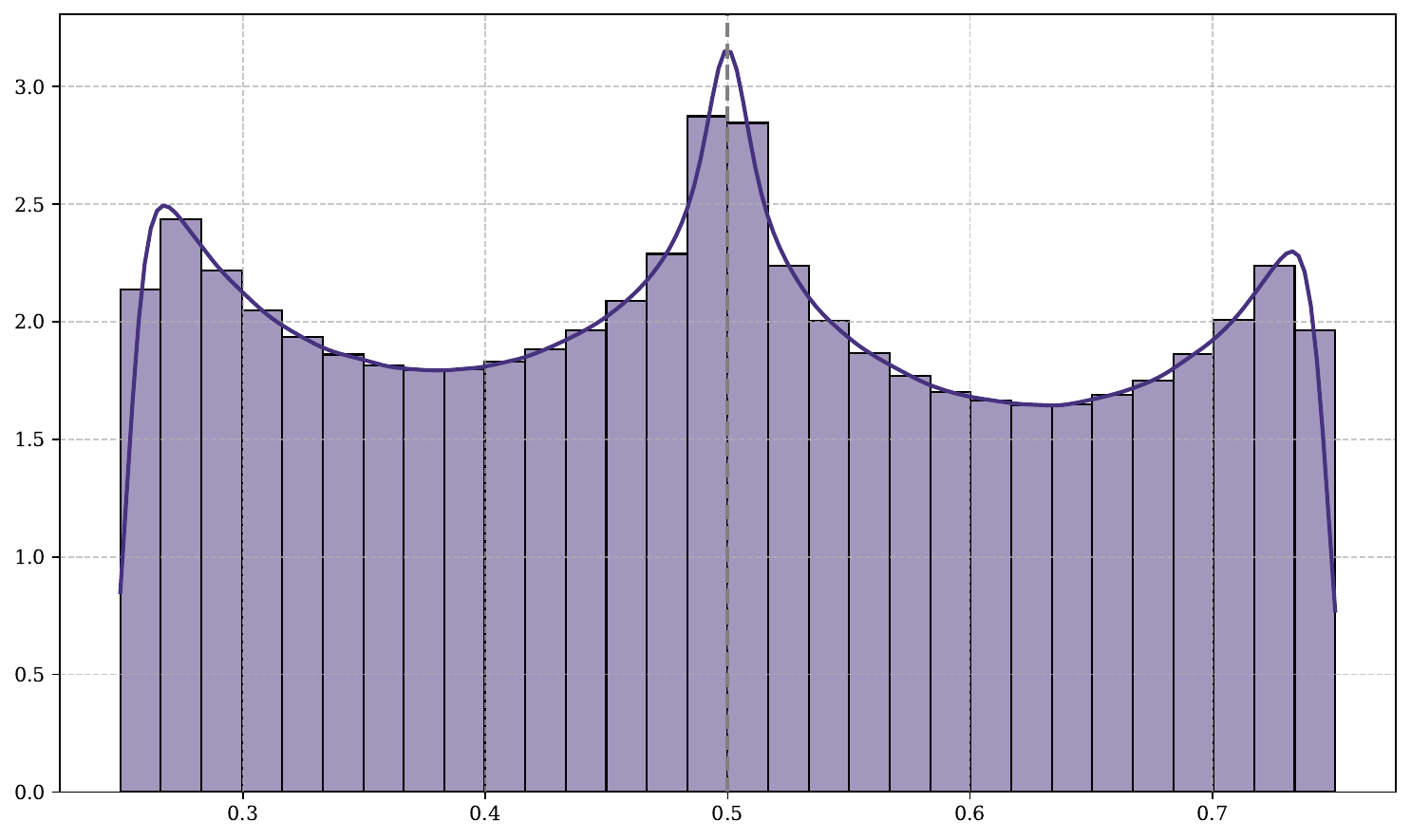}
    \caption{Kernel density estimation and frequencies over values. Distribution of the agreeing probabilities $\Prob\left[g_{U,i}g_{C,i} > 0 \,\big|\, \hat g_{U}, \hat g_{C}\right]$ for all the parameters of a VGG16 during an unlearning epoch step. Doted vertical line at $p = \nicefrac{1}{2}$.}
    \Description{Distribution of the agreeing probability over all the $N=38784$ parameters of a ResNet18 during an unlearning epoch step}
    \label{fig:distrprob}
\end{figure}
Figure~\ref{fig:distrprob} shows the empirical distribution of the probability~\eqref{eq:AgreeProb} computed over all the components $\theta_i$ for a given batch. Roughly speaking, there is one quarter of the parameters where the signs of the true gradient components likely disagree (small probability), one quarter where they likely agree (large probability), and one half where nothing is certain (probability around \nicefrac{1}{2}). This latter case happens typically when at least one of $\hat g_{U,i}$ or $\hat g_{C,i}$ takes a small value compared to the associated standard deviation, so that $\phi_{U,i}$ or $\phi_{C,i}$ is close to $\nicefrac{1}{2}$, and so is the probability~\eqref{eq:AgreeProb}. The figure shows that, conditioned on the observation of a batch, those different regimes are evenly represented.

An implementation issue is the evaluation of the standard deviation of the estimation noises $N_{U,i}$ and $N_{C,i}$.
A simple solution is to retrieve the standard deviation of the $i$-th component of the gradient per datum included in the batch.
This quantity is indeed computed efficiently as the stacked moving average second-order momentum in the PyTorch implementation of the Adam optimizer~\cite{kingma2014adam}.

\subsection{New Masking Strategy}
\label{sec:designmask}
In this section, the update is applied on the empirical gradients, but leveraging on Prop.~\ref{prop:diragreeprob} for designing new masks dedicated to batch processing.
A first idea is as follows:
\begin{align} 
    \Delta_{\PROB}&\doteq - m_{\PROB}\odot \sif \left(\hat g_{U}, \hat g_{C}\right)\quad\text{with}\label{eq:MethodPROB}\\
    m_{\PROB}(\hat g_{U}, \hat g_{C})_i &\doteq
    \begin{cases}
        1 & \text{if}\,\, \Prob\left[g_{U,i}g_{C,i}>0  \,\big|\,  \hat g_{U}, \hat g_{C}\right]>p, \\
        0 & \text{otherwise}.
    \end{cases}
    \label{prob-selection}
\end{align}
where $p\in(0,1)$ is a level of certainty.
In a nutshell, we modify a parameter if and only if we are certain that the update is locally feasible (see Fig.~\ref{fig:maskingSalUn}).

The novelty over saliency masking is threefold. First, SalUn selects parameters where the gradient ascent is confident but implicitly wrongly assumes that all parameter variances are equal. By introducing variance, we can achieve more precision. Second, we replace the arbitrary saliency threshold parameter with an interpretable quantile. 
Third but not least, SalUn requires the computation of a third gradient to compute the masking while our techniques recycle the already computed gradients.

\begin{proposition}[Properties of $m_{\PROB}$]
    \label{prop:Properties}
    The mask $m_{\PROB}$~\eqref{prob-selection} equals the mask $m_\text{AND}$~\eqref{eq:Mask} applied on the empirical gradients at a given batch in the following cases:
    \begin{enumerate}
        \item The level of certainty is set to $p=\nicefrac{1}{2}$,
        \item All variances $\Sigma_{U,(i,i)}$ and $\Sigma_{C,(i,i)}$ tend to 0 and $p\in(0,1)$.
    \end{enumerate}
\end{proposition}

\begin{figure}[tbh]
    \centering
    \includegraphics[width=\rat\columnwidth]{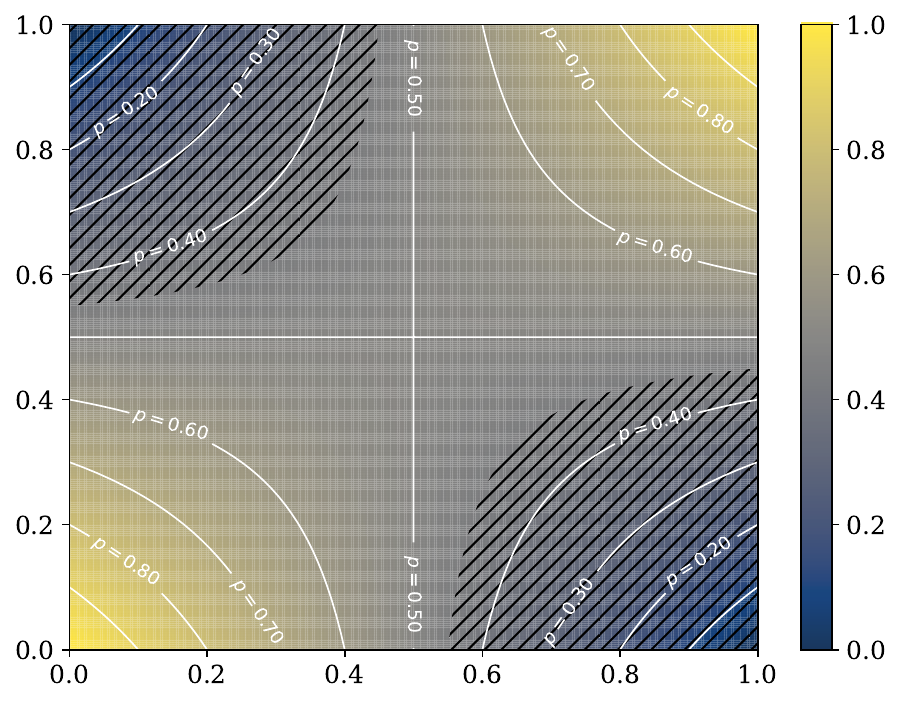}
    \caption{Heatmap of the probability $\Prob\left[g_{U,i}g_{C,i}>0  \,\big|\,  \hat g_{U}, \hat g_{C}\right]$ with respect to coordinates $(\phi_{U,i}, \phi_{C,i})\in(0,1)^2$. Hatched areas correspond to a probability lower than $p$, provoking a masking by~\eqref{prob-selection} - $p=0.45$ in this example.}
    \label{fig:heatmap}
    \Description{Heatmap of the probability}
\end{figure}

\subsection{The Focus Vector}
\label{sec:focusvect}
The level of certainty $p$ is a hyperparameter whose setting is ad hoc. An alternative is a probabilistic mask where each component is randomly sampled according to a Bernoulli distribution $\ber$: 
\begin{align} 
    \Delta_{\BER}&\doteq - m_{\BER}\odot \sif\left(\hat g_{U}, \hat g_{C}\right)\quad\text{with}\\
    m_{\BER,i}~&\sim \ber(\focus_{i})\quad\text{where}\, 
    \focus_{i} \doteq \Prob\left[g_{U,i}g_{C,i}>0  \,\big|\,  \hat g_{U}, \hat g_{C}\right].
\end{align}
This probabilistic selection leads, in expectation over $(\hat{g}_U,\hat{g}_C)$, to a weighting of the update acting as a focus on the parameters more likely to contribute to solving the optimization problem~\eqref{problem}:
\begin{equation}
    \begin{split}
        \Exp\left[\Delta_{\BER} \,\big|\, \hat g_U, \hat g_C\right] 
        &= -\Exp\left[m_{\BER}\odot \sif \left(\hat g_U, \hat g_C\right) \,\big|\, \hat g_U, \hat g_C\right]\\
        &= -\focus\odot \sif\left(\hat g_U, \hat g_C\right),
    \end{split}
\end{equation}
with $\focus\doteq(\focus_i)_i$ the \textit{focus vector}. Again, by the proof of Prop.~\ref{prop:Properties} Case 2, this tends almost surely, when the variance goes to zero, to the update~\eqref{eq:Aggreg} applied to the batch gradients. 
By continuity for low standard deviations, one understands that this update is a good direction for decreasing both $U$ and $C$.

It explains our final design for the update:
\begin{equation}
    \Delta_{\FOCUS} \doteq -\focus\odot \sif\left(\hat g_U, \hat g_C\right).
    \label{eq:MethodFOCUS}
\end{equation}

\input{main/5_algo}
Knowing that the scalar product of two random vectors concentrates around zero in high-dimensional space ~\cite{cho2009inner}, we state a feasible theoretical guarantee for this last version of the update.

\begin{proposition}[Focus update's Theoretical Guarantee]
    Let the aggregation $\sif$ be the linear positive combination~\eqref{eq:linG} of the gradients with $\alpha, \beta > 0$.
    Suppose that the measured gradients $\hat g_U$ and $\hat g_C$ are in expectation scalarly independent: 
        $\mathbb{E}\left[\langle \hat g_U, \hat g_C \rangle\right] = 0$.
    Then, the update $\Delta_{\FOCUS} = - \focus \odot \left(\alpha \hat g_U + \beta \hat g_C\right)$ is feasible in expectation, and for any sufficiently small learning rate $\eta > 0$:
    \begin{align}
        \mathbb{E}&\left[U(\theta_0+\eta\Delta_\FOCUS(\theta_0))\right] \leq U(\theta_0) \\
        \mathbb{E}&\left[C(\theta_0+\eta\Delta_\FOCUS(\theta_0))\right] \leq 0.
    \end{align}
    \label{prop:focusguarantee}
\end{proposition}
We verified experimentally the hypothesis $\mathbb{E}\left[\langle \hat g_U, \hat g_C \rangle\right] = 0$. The empirical values of $\langle \hat g_U, \hat g_C \rangle$ are initially about $\pm 10^{-5}$ and continue decreasing down to $\pm 10^{-10}$ during the unlearning procedure.

\begin{proposition}[Properties of the Focus Vector $\focus$]
    \label{prop:PropertiesFocus}
    The term-by-term multiplication by the focus vector $\focus$ corresponds to 
    \begin{enumerate}
        \item the mask $m_\text{AND}$~\eqref{eq:Mask} applied on the empirical gradients when the variances $\Sigma_{U,(i,i)}$ and $\Sigma_{C,(i,i)}$ tend to $0$, $\forall i$,
        \item halving the learning rate when the variances $\Sigma_{U,(i,i)}$ and $\Sigma_{C,(i,i)}$ tend to $+\infty$, $\forall i$.
    \end{enumerate}
\end{proposition}

Prop.~\ref{prop:PropertiesFocus} shows that the focus vector add-on is a soft trade-off between the $\AND$ mask when the gradient estimation noise is small and a slowdown of the learning rate when the variance is too high.

%% file: main/5_algo.tex
\begin{algorithm}[tb]
\caption{\textsc{Focus Vector for Unlearning }}
\label{alg:focus-unlearning}
\begin{algorithmic}[1]

\Require Objective and constraint dataset and loss $\mathcal{D}_U$, $\mathcal{D}_C$ and $\loss_U$, $\loss_C$; batch size $B$; aggregation function $\sif$; initial parameters $\theta_u \gets \theta_0$; learning rate $\eta$; stability term $\epsilon$;

\Ensure Updated model parameters $\theta_u$ after unlearning

\For{$e = 1, 2, \dots$} \Comment{\textcolor{teal}{Epoch loop}}
    \For{$b = 1, 2, \dots$} \Comment{\textcolor{teal}{Batch loop}}
    
        \State \textcolor{teal}{Sample batches} 
        \State \hspace{\algorithmicindent} $(x^U_i, y^U_i)_{i=1}^B \sim \mathcal{D}_U$, $(x^C_i, y^C_i)_{i=1}^B \sim \mathcal{D}_C$
        
        \State \textcolor{teal}{Compute gradients}
        \State \hspace{\algorithmicindent} $\hat{g}_U \gets \frac{1}{B} \sum_{i=1}^B \nabla L_U(M_{\theta_u}(x^U_i), y^U_i)$
        \State \hspace{\algorithmicindent} $\hat{g}_C \gets \frac{1}{B} \sum_{i=1}^B \nabla L_C(M_{\theta_u}(x^C_i), y^C_i)$
        
        \State \textcolor{teal}{Retrieve second-order statistics}
        \State \hspace{\algorithmicindent} $\Sigma_U \gets \texttt{AdamMoments}(L_U)$
        \State \hspace{\algorithmicindent} $\Sigma_C \gets \texttt{AdamMoments}(L_C)$
        
        \State \textcolor{teal}{Compute per-parameter focus}
        \State \hspace{\algorithmicindent} $\phi_U \gets \Phi\left( \nicefrac{\hat{g}_U}{\sqrt{\mathrm{diag}(\Sigma_U) + \epsilon}} \right)$
        \State \hspace{\algorithmicindent} $\phi_C \gets \Phi\left( \nicefrac{\hat{g}_C}{\sqrt{\mathrm{diag}(\Sigma_C) + \epsilon}} \right)$
        \State \hspace{\algorithmicindent} $\focus \gets \phi_U \odot \phi_C + (1 - \phi_U) \odot (1 - \phi_C)$
        
        \State \textcolor{teal}{Compute update direction}
        \State \hspace{\algorithmicindent} $\Delta_\FOCUS \gets -\focus \odot \sif \left(\hat{g}_U, \hat{g}_C\right)$
        
        \State \textcolor{teal}{Update parameters}
        \State \hspace{\algorithmicindent} $\theta_u \gets \texttt{Optimize}(\theta_u, \Delta_\FOCUS, \eta)$
        
    \EndFor
\EndFor
\Statex \Return $\theta_u$
\end{algorithmic}
\end{algorithm}

%% file: main/tables/table_sub_1.tex
\begin{table}[tbh]
  \caption{Comparison of mask-based add-ons after 10 epochs of 10\% random data forgetting. Upper part: SRL - Cifar10 - VGG16. Bottom part: NGPlus - SVHN - ResNet18.}
  \label{tab:masks}
  \centering
  \resizebox{\textwidth}{!}{
  \begin{tabular}{lcccccc}
    \toprule
Methods & \textbf{MIA entropy} & \textbf{rUA} & \textbf{TA} & \textbf{RA} & \textbf{UA} & \textbf{FID} \\
    \midrule
SRL &  0.89 $\pm$ 0.01 & 4.62 $\pm$ 0.87 & 92.77 $\pm$ 0.12 & 99.24 $\pm$ 0.16 & 97.78 $\pm$ 0.39 & 93.21 $\pm$ 0.40 \\
SRL - SalUn &  0.89 $\pm$ 0.01 & 4.70 $\pm$ 0.78 & 92.80 $\pm$ 0.14 & 99.25 $\pm$ 0.19 & 97.85 $\pm$ 0.32 & 93.36 $\pm$ 0.41 \\

\cmidrule(lr){2-7}

SRL - $\AND$ &  \textbf{0.69 $\pm$ 0.01} & 2.30 $\pm$ 0.47 & 92.18 $\pm$ 0.08 & 98.76 $\pm$ 0.21 & 95.45 $\pm$ 0.52 & 91.81 $\pm$ 0.41 \\
SRL - $\PROB$ & \textbf{0.76 $\pm$ 0.02} & \textbf{0.45 $\pm$ 0.36} & 91.69 $\pm$ 0.16 & 98.71 $\pm$ 0.22 & 93.61 $\pm$ 0.74 & 90.74 $\pm$ 0.47 \\
\textbf{SRL - $\FOCUS$} &  \textbf{0.77 $\pm$ 0.03} & \textbf{0.41 $\pm$ 0.50} & 91.71 $\pm$ 0.15 & 98.75 $\pm$ 0.23 & 93.57 $\pm$ 0.34 & 90.60 $\pm$ 0.14 \\

    \midrule
NGPlus &  0.94 $\pm$ 0.01 & 3.13 $\pm$ 0.76 & 94.75 $\pm$ 0.54 & 99.74 $\pm$ 0.06 & 97.98 $\pm$ 0.77 & 94.24 $\pm$ 0.74 \\
NGPlus - SalUn &  0.93 $\pm$ 0.02 & 2.67 $\pm$ 1.37 & 94.53 $\pm$ 0.91 & 99.73 $\pm$ 0.01 & 97.53 $\pm$ 1.42 & 93.90 $\pm$ 1.20 \\

\cmidrule(lr){2-7}

NGPlus - $\AND$ & \textbf{0.54 $\pm$ 0.02} & -46.14 $\pm$ 7.45 & 75.99 $\pm$ 3.39 & 88.57 $\pm$ 2.96 & 48.67 $\pm$ 7.72 & 47.67 $\pm$ 7.91 \\
NGPlus - $\PROB$ &  \textbf{0.88 $\pm$ 0.03} & \textbf{0.43 $\pm$ 1.34} & 93.18 $\pm$ 1.29 & 99.53 $\pm$ 0.11 & 95.24 $\pm$ 1.55 & 92.24 $\pm$ 1.75 \\
\textbf{NGPlus - $\FOCUS$} & \textbf{0.89 $\pm$ 0.02} & \textbf{0.14 $\pm$ 0.80} & 93.42 $\pm$ 0.74 & 99.56 $\pm$ 0.04 & 94.99 $\pm$ 0.94 & 92.25 $\pm$ 1.06 \\

    \bottomrule
  \end{tabular}
  }
\end{table}

%% file: main/6_experiments.tex
\subsection{Experimental Setup}
\label{sec:expsetup}
All the experiments code is available at \url{https://github.com/owl1996/UnlearningFocusVector} and Sect.~\ref{sec:results} only presents snapshots of this extensive experimental body. 

\paragraph{Architectures and Datasets.} Our experimental protocol focuses on image classification tasks with the datasets CIFAR-10~\cite{krizhevsky2009learning} and SVHN~\cite{netzer2011reading}, and the deep neural network architectures ResNet18 model~\cite{he2016deep} and VGG16~\cite{simonyan2014very}.

\paragraph{Unlearning Scenarios.} The protocol  considers two scenarios about the dataset $\sD_\FORGET$ to be forgotten:
\begin{itemize}
    \item $\sD_\FORGET$ is a random fraction of the whole training data,
    \item $\sD_\FORGET$ is a random fraction of a specific class.
\end{itemize}
For the first scenario, random forgetting, $\sD_\FORGET$ represents 5\%, 10\%, or 50\% of the entire training data.
For the second scenario, in class forgetting, $\sD_\FORGET$ represents 10\%, 40\%, or 75\% of a class data.

\paragraph{Unlearning Methods} 
Saliency techniques and our updating procedures $\AND$~\eqref{eq:Aggreg}, $\PROB$~\eqref{eq:MethodPROB}, and $\FOCUS$~\eqref{eq:MethodFOCUS} are coined 
\emph{add-ons} because they are generics and they can be applied to different unlearning methods (defined by their losses, constraints, and datasets) introduced in Sect.~\ref{sec:GenFrame}.
Our experimental protocol considers SRL, NGPlus, and SCRUB.

\paragraph{Benchmark}
For each method, a baseline is the ideal model trained from scratch in the same way as the initial model but without the forget set $\sD_\FORGET$.
We consider the baselines Mask-Small-Gradients (MSG)~\cite{msg} and 
Convolution-Transpose (CT)~\cite{ct}, in addition to the following cited in Sect.~\ref{sec:RelWorkUnlearn}:
Fine-Tuning (FT)~\cite{Warnecke2021MachineUO},
SalUn~\cite{Fan2023SalUnEM} and SCRUB~\cite{kurmanji2023towards}.
Of note, MSG and CT participated in the NeurIPS'2023 Machine Unlearning competition~\cite{NeurIPSChallenge}. Moreover, FT, MSG and CT are evaluated as the strongest baselines in~\cite{cadet2024deep} due to their consistent performances across different datasets. In particular, MSG and CT show robustness against powerful per-sample MIAs like U-LiRA. SalUn is similar to our approach as it masks the least salient weights of the gradient ascent over the forget set. It follows the same spirit of our add-ons and is thus a competitor.

\paragraph{Hyper Parameters}
The batch size is 256 for any dataloader, both for training and unlearning.
We train the initial and the exact ideal models for $100$ epochs with
an initial learning rate set to $10^{-1}$.
For unlearning, the initial learning rate $\eta$ equals $10^{-4}$.
The linear combination~\eqref{eq:linG} takes $\alpha = 0.05$ for the gradient of loss $L_U$ and $\beta = 0.95$ for the gradient of loss $L_C$.
By default, the $\PROB$ add-on sets the quantile parameter $p = 0.3$. SalUn~\cite{Fan2023SalUnEM} uses a median absolute value mask; for SCRUB, we let the additional loss with ponderation $\gamma = 1$ (see Sect.~\ref{sec:GenFrame}).

\paragraph{Multiple Runs}
For each given setting, the initial model, the unlearning procedure, and the exact ideal model are computed several times with different random seeds to measure means and standard deviations reported in the figures and tables.

\subsection{Evaluation Metrics.}
\label{sec:EvaluationMetrics}
Our evaluation essentially follows the criteria outlined in \cite{jia2023model} as discussed in Sect.~\ref{sec:EvalMetrics}:
\begin{itemize}
    \item RTE: Run Time Efficiency (in seconds),
    \item MIA: Membership Inference Attack,
    \item TA: Test Accuracy,
    \item RA: Retain Accuracy,
    \item UA: Unlearning Accuracy.
\end{itemize}
Our protocol train five different support vector classifiers as decision rules $b$ for the MIA~\eqref{eq:MIADef}, each exploiting different data~\cite{hu2022membership, yeom2018privacy, li2022leaks, bertran2023scalable, choquette2021label}. The different training features used for the MIA are as follows: 
\textit{correctness} (binary variable indicating whether the model well classifies the input data),
\textit{confidence} (probability score of the assigned class),
\textit{logits} (the logits output by the model),
\textit{entropy} (the entropy of the output of the model),
\textit{mix entropy} (the mix entropy of the logits). 
For simplicity, we do not show all five MIA metrics, but report the performance of the ones that achieve the best average score across comparisons: entropy and mix entropy.

We introduce two other metrics for the unlearning evaluation.
Most works consider that a low the UA is a good indicator but it forces the unlearned model to misclassify the data to be forgotten.
The problem is that these data are regular testing data for an ideal model $M_{\IdealUModel}$ which may still classify them correctly.
We define the Relative Unlearning Accuracy (rUA) as:
\begin{equation}
    \mathrm{rUA} \doteq \Prob_{(x,y)\sim \sD_\FORGET}\left[ M(x) = y\right] - \Prob_{(x,y)\sim \sD_\FORGET}\left[M_{\theta^{\star}} (x) = y \right].
    \label{eq:rUA}
\end{equation}
A good forgetting of the data expects rUA to be close to zero. When it is strongly positive, the unlearned model still classifies the data to forget too well, while the ideal model may have a lower accuracy over these data. When it is strongly negative, the procedure is too brutal, provoking catastrophic forgetting of the model. We define the Fidelity score (FID) (\textit{a.k.a.} consistency~\cite{xu2023machineunlearningsurvey}):
\begin{equation}
    \mathrm{FID} \doteq \mathbb{P}_{(x,y)\sim D_\FORGET}\left[ M(x) = M_{\theta^\star} (x) \right].
    \label{eq:Fid}
\end{equation}
It measures how well the unlearned model mimics the behavior of the ideal model. When the ideal model fails because too much data was forgotten, entailing a loss of the model's utility, this metric gauges whether the unlearned model makes the same mistakes.

These new metrics are applicable only to laboratory experiments, in the sense that they cannot be directly observed in practice, as the ideal model is never fully realized. Yet, they offer a better understanding of what is expected from an unlearning.

%% file: main/6_results.tex

\subsection{Results}
\label{sec:results}

According to our study, when keeping the unlearning method and the scenario fixed (class-wise or random), the change of data or model architecture has no impact: the order of the add-ons techniques across all the metrics is conserved. Then, to avoid redundancy and excessive figures, we only report the non-intersecting part of the configurations to exhibit the different regimes of unlearning on which we test our add-ons.

\subsubsection{Comparison of Masks}
Table~\ref{tab:masks} reports the metrics of mask-based add-ons for two unlearning methods, SRL and NGPlus.
It includes no masking (baseline version of SRL and NGPlus), SalUn~\cite{Fan2023SalUnEM}, and our variants $\AND$~\eqref{eq:Aggreg}, $\PROB$~\eqref{eq:MethodPROB}, and $\FOCUS$~\eqref{eq:MethodFOCUS}.
Our variants bring better MIA and rUA metrics for a moderate price to pay in terms of TA, RA, and FID.
One exception is NGPlus-$\AND$, which does not perform well.
It illustrates the drawback of this add-on when applied to batch processing:
an agreement on the sign of the noisy batch gradient components may not lead to such an agreement on the true gradient.
Hence, the mask $\AND$ may select parameters incompatible with a locally feasible update. 
The add-ons $\PROB$ and $\FOCUS$ correct this erroneous behavior.

\begin{figure*}[tb]
    \centering
    \includegraphics[width = \rat\linewidth]{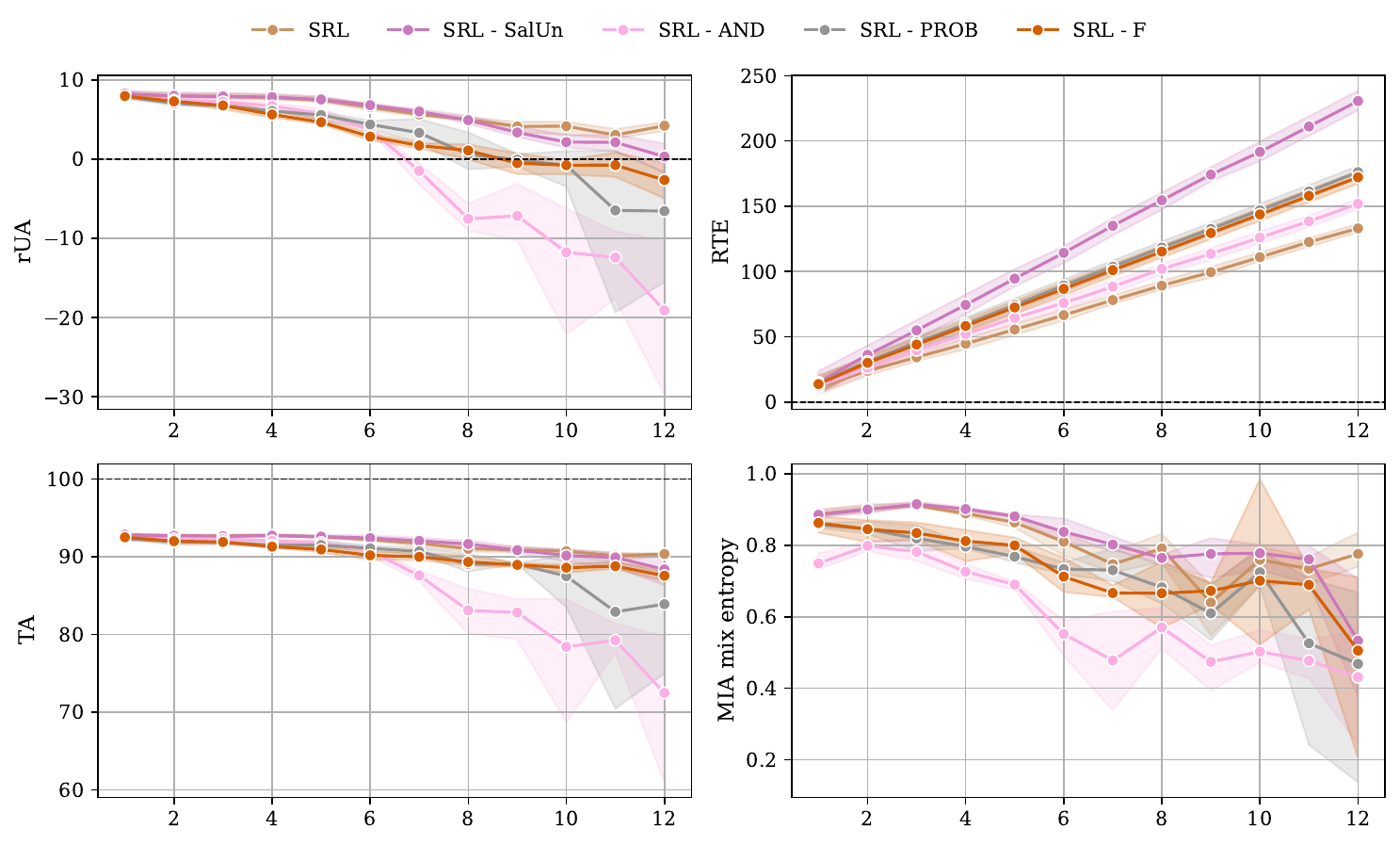}
    \caption{Comparison of add-ons (SalUn, $\AND$, $\PROB$ and $\FOCUS$) over SRL method. Evolution of the metrics over the number of epochs. Cifar10 -- VGG16 -- 50\% Random forgetting}
    \label{fig:addonsSRL}
    \Description{Comparison of Baselines}
\end{figure*}

This phenomenon is also visible in Fig.~\ref{fig:addonsSRL} comparing the add-ons over SRL.
The simple $\AND$ add-on is more unstable, while $\FOCUS$ and $\PROB$ add-ons are a good trade-off between efficiency and efficacy.
In each case, the three add-ons we introduce are those that lower the MIA and absolute rUA with few steps of unlearning.

As for RTE, our add-on is $\approx 22\%$ slower than the baseline method it is applied to.
Of note, SalUn~\cite{Fan2023SalUnEM} masks the parameters with the least salient weights of the gradient ascent over the forget set.
It requires the computation of a third gradient.
This is already available in the NGPlus method, so for this particular trivial method SalUn is as fast as our add-on.
Yet, it is not the case in SRL and SCRUB methods, for which SalUn is $\approx 55\%$ slower than the baseline.

\subsubsection{Generalization w.r.t. the Unlearning Methods}
\begin{figure*}[tb]
    \centering
    \includegraphics[width=\siz\linewidth]{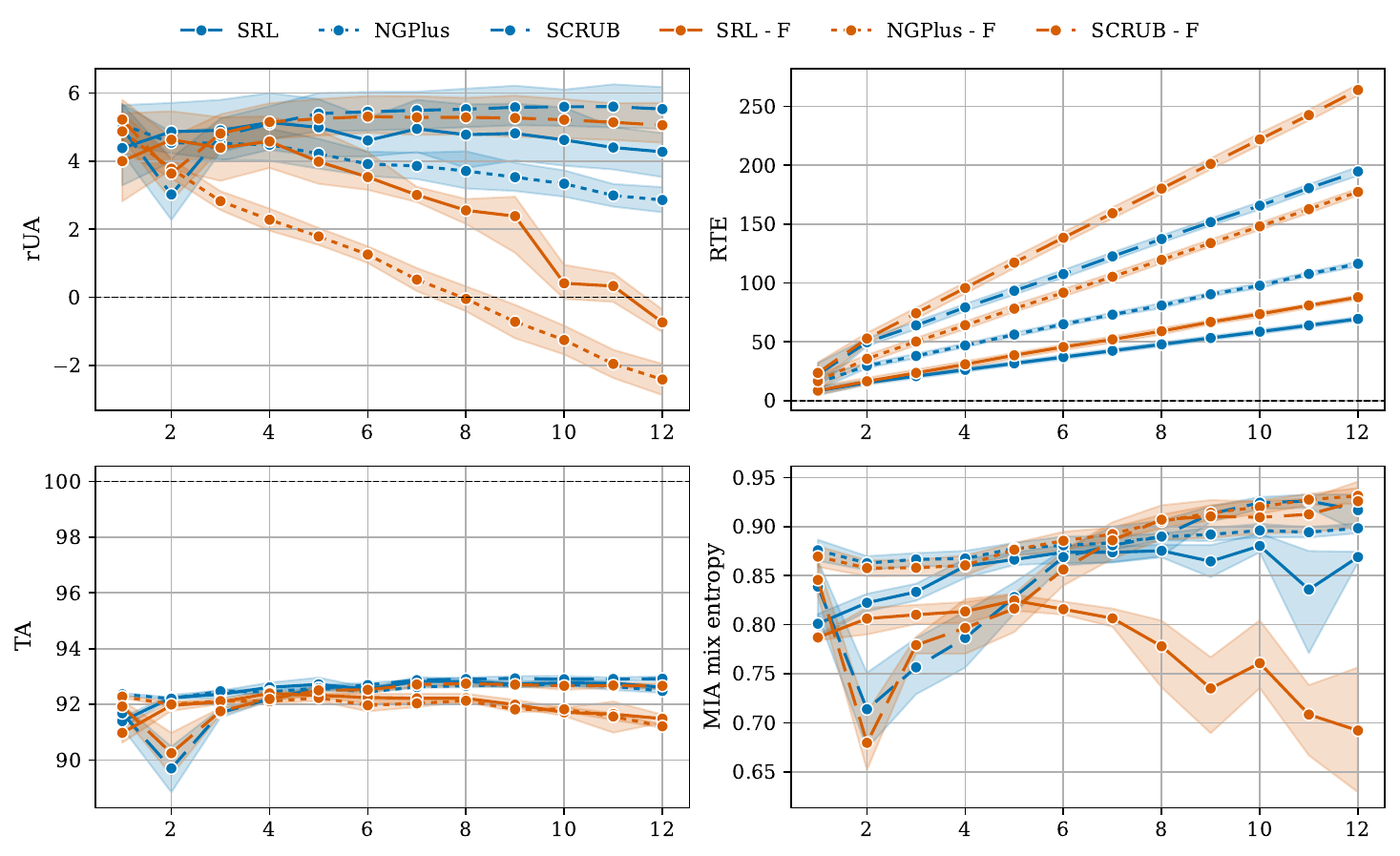}
    \caption{Benefit of the focus vector add-on $\FOCUS$~\eqref{eq:MethodFOCUS}. Evolution of the metrics over the number of epochs for three methods (SRL, NGPlus, and SCRUB). Cifar10 -- VGG16 --  10\% Class-wise forgetting}
    \label{fig:addon-methods}
    \Description{Adding focus vector method}
\end{figure*}
Fig.~\ref{fig:addon-methods} shows the main benefit of the focus vector add-on $\FOCUS$~\eqref{eq:MethodFOCUS}: 
The rUA goes more rapidly down to zero, no matter the method we plug it into (NGPlus, SRL, or SCRUB).
As for the MIA, the focus vector add-on is always under the initial method score curve.
The price to pay is moderate: a slight loss on the TA and a small increase on the RTE.

\subsubsection{Differences between SRL, SCRUB, and  NGPlus}
\begin{figure*}[tb]
    \centering
    \includegraphics[width=\siz\linewidth]{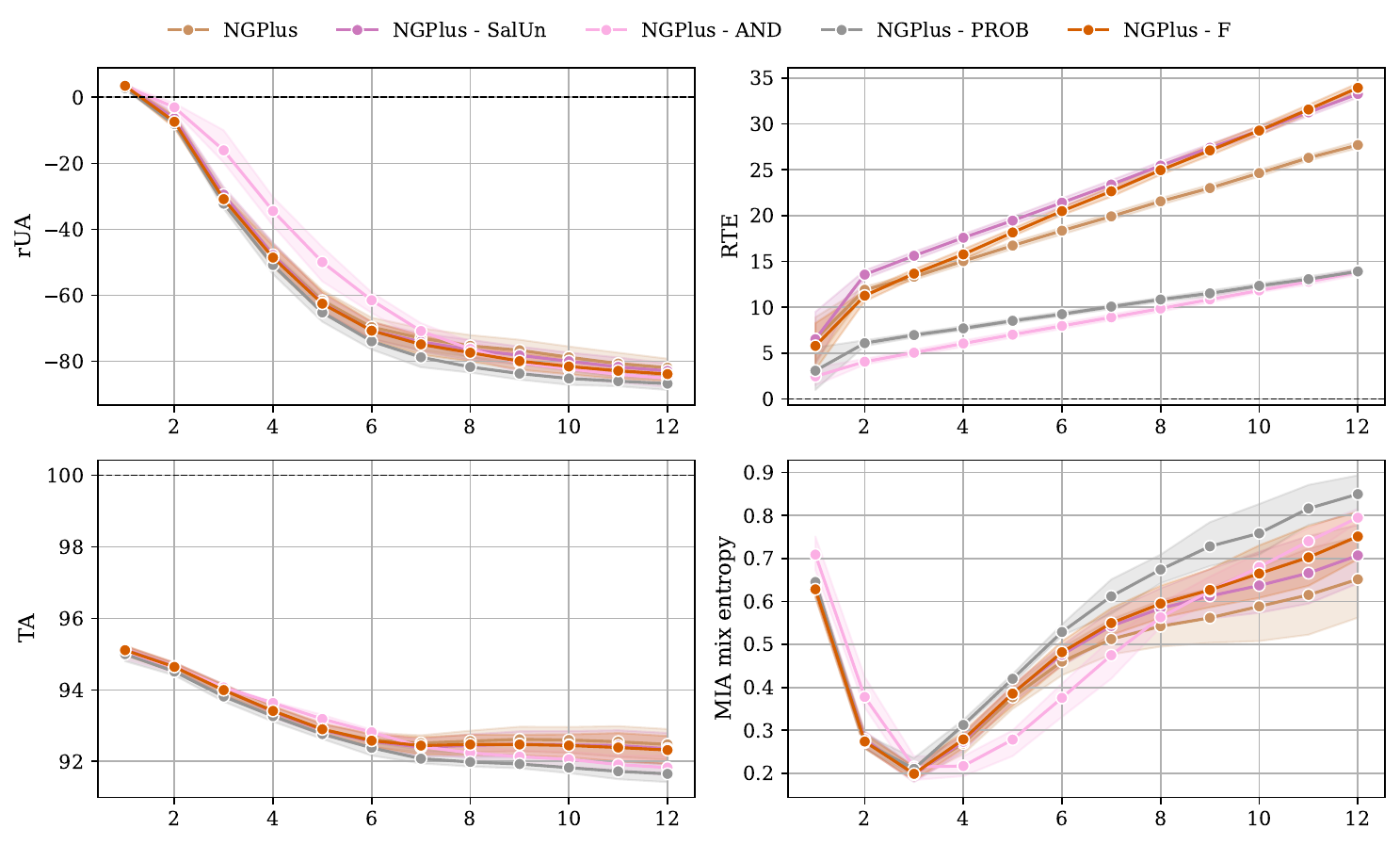}
    \caption{Comparison of add-ons (SalUn, $\AND$, $\PROB$ and $\FOCUS$) over NGPlus method. Evolution of the metrics over the number of epochs. NGPlus -- SVHN -- ResNet18 -- 10\% Class-wise forgetting}
    \label{fig:addonsngplus}
    \Description{Comparison of Baselines}
\end{figure*}
Across the multiple settings of our experimental benchmark, we notice that NGPlus may have a behavior different than SRL and SCRUB.
Fig.~\ref{fig:addonsngplus} shows a class-wise forgetting setup where all add-ons lead to similar results.
One or two steps of unlearning is sufficient here and indeed recommended to avoid destroying the accuracy of the specific class due to the gradient ascent, which results in stronger MIA. 

\subsubsection{Benchmark}

\input{main/tables/table_benchmark}

Table~\ref{tab:benchmark} overall shows that our add-ons are competitive in unlearning efficacy compared to the baselines.
The MIA performs poorly over the ideal model, but we are interested in whether the add-on lowers the MIA's performance.
Indeed, our add-ons always lower the initial unlearning method's score.
MSG and CT look resilient to MIA at early epochs because they consist in initially jeopardized a portion of the model parameters.

\begin{figure*}[tb]
    \centering
    \includegraphics[width=\siz\linewidth]{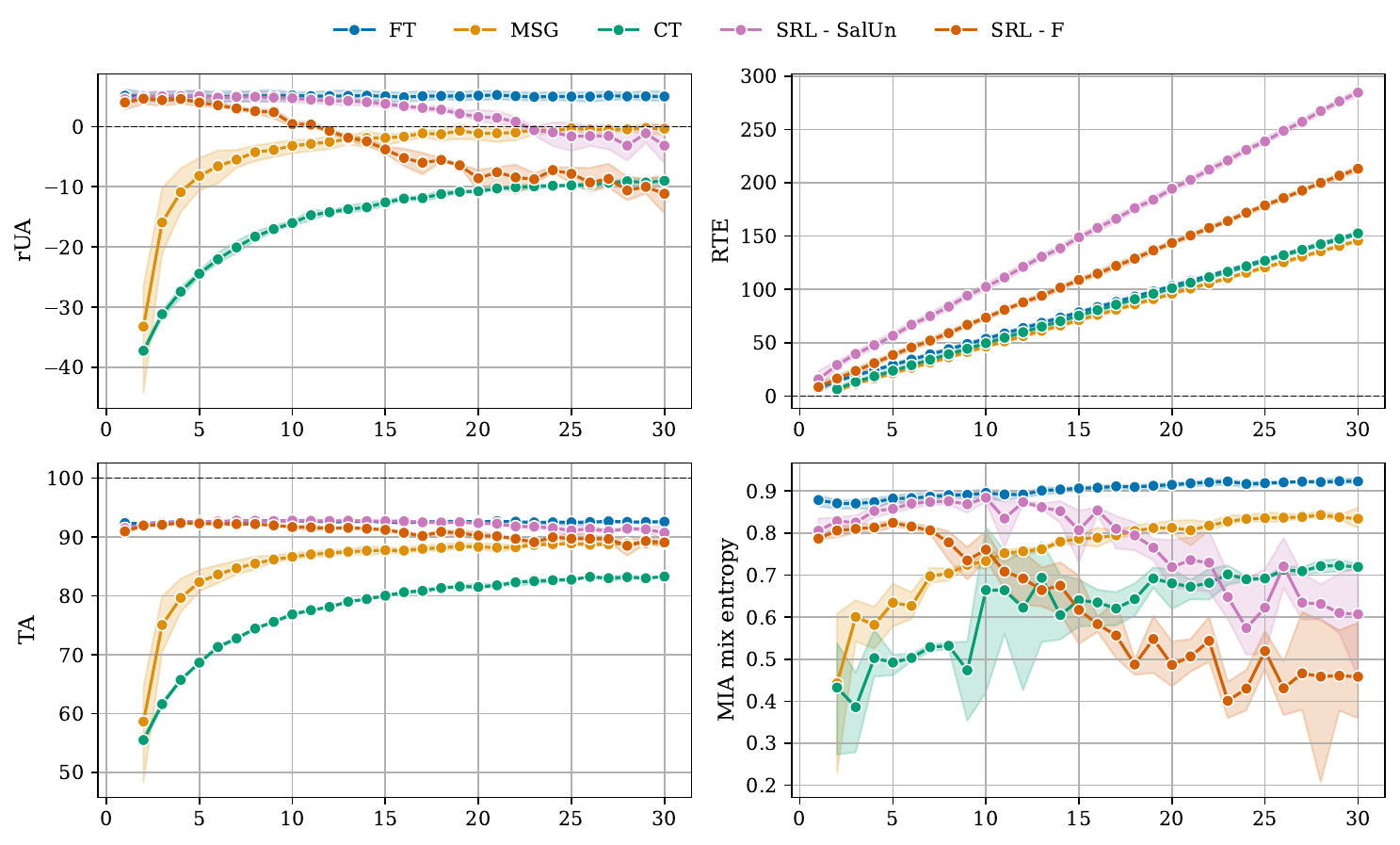}
    \caption{Comparison of baselines with add-on $\FOCUS$ over SRL. Evolution of the metrics over the number of epochs. Cifar10 -- VGG16 --  10\% Random forgetting}
    \label{fig:bestimage}
    \Description{Comparison of Baselines}
\end{figure*}

\begin{figure*}[tb]
    \centering
    \includegraphics[width=\siz\linewidth]{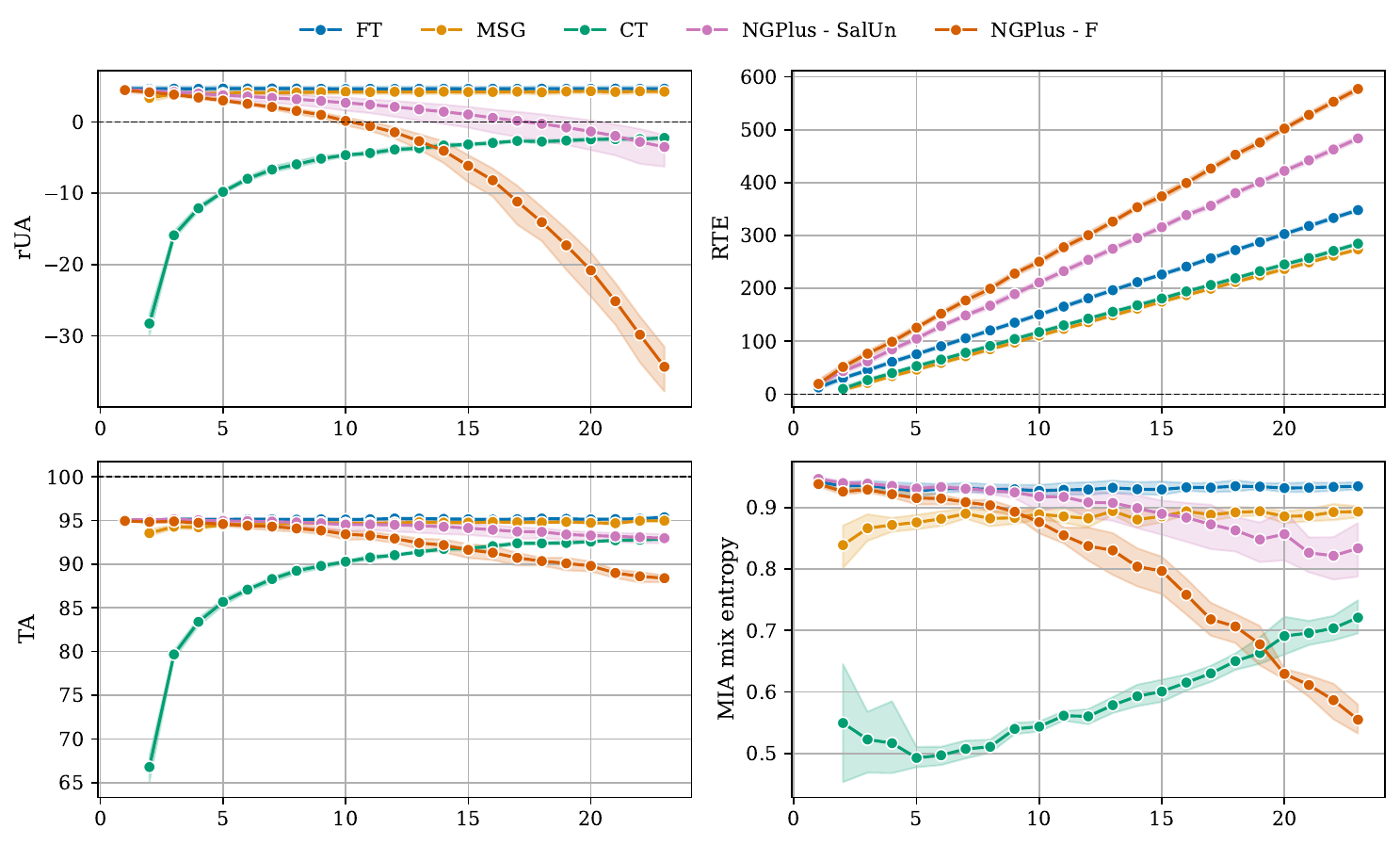}
    \caption{Comparison of baselines with add-on $\FOCUS$ over NGPlus. Evolution of the metrics over the number of epochs. SVHN -- ResNet18 -- 10\% Random forgetting}
    \label{fig:bestimagengplus}
    \Description{Comparison of Baselines}
\end{figure*}

Yet, this initial modification also pulls down the other metrics rUA and TA as shown in Fig.~\ref{fig:bestimage} and~\ref{fig:bestimagengplus}. Those figures show the evolution of the metrics along the epochs, enabling us to compare the efficacy and efficiency of the methods.
The remaining epochs try to recover this parameter randomization in CT and MSG.
This slowly improves rUA and TA but also favors the MIA.
On the contrary, our solutions converge faster to a lower rUA. Also, the MIA is less efficient.
We noticed that MSG needs a more careful setting of the threshold across the different architectures than the 30\% proposed by the authors.
Our solutions are always faster at decreasing the absolute value of rUA while stabilizing the test accuracy right from the first epochs.

To summarize, the relative unlearning accuracy metric quickly reaches zero with our solutions.
The competitors need about twice as much time to reach this level of rUA and comparable accuracies.
It shows that, in practice, our solutions improve efficiency.

%% file: main/tables/table_benchmark.tex
\begin{table}[thb]

  \caption{Benchmark - Epoch 10 - Cifar10 - VGG16 - 10\% Random Forgetting}
  \label{tab:benchmark}
  \centering
  \resizebox{\textwidth}{!}{
  \begin{tabular}{lcccccc}
    \toprule
Methods &  \textbf{MIA entropy} & \textbf{rUA} & \textbf{TA} & \textbf{RA} & \textbf{UA} & \textbf{FID} \\
\midrule
Initial  & 0.91 $\pm$ 0.01 & 5.82 $\pm$ 0.84 & 92.66 $\pm$ 0.10 & 98.63 $\pm$ 0.27 & 98.73 $\pm$ 0.27 & 93.00 $\pm$ 0.56 \\
Ideal & 0.84 $\pm$ 0.01 & \textbf{0.00 $\pm$ 0.00} & 92.53 $\pm$ 0.16 & 99.56 $\pm$ 0.05 & 92.91 $\pm$ 0.71 & 100.0 $\pm$ 0.00 \\

\cmidrule(lr){2-7}
FT & 0.91 $\pm$ 0.01 & 5.15 $\pm$ 0.88 & 92.75 $\pm$ 0.15 & 99.18 $\pm$ 0.14 & 98.3 $\pm$ 0.39 & 93.44 $\pm$ 0.52 \\
MSG & \textbf{0.76 $\pm$ 0.02} & -3.20 $\pm$ 1.13 & 86.65 $\pm$ 0.56 & 93.07 $\pm$ 0.84 & 89.96 $\pm$ 0.66 & 88.42 $\pm$ 0.69 \\
CT &  \textbf{0.65 $\pm$ 0.08} & -16.01 $\pm$ 0.37 & 76.87 $\pm$ 0.33 & 80.45 $\pm$ 0.27 & 77.14 $\pm$ 0.37 & 77.51 $\pm$ 0.53 \\
SRL &  0.89 $\pm$ 0.01 & 4.62 $\pm$ 0.87 & 92.77 $\pm$ 0.12 & 99.24 $\pm$ 0.16 & 97.78 $\pm$ 0.39 & 93.21 $\pm$ 0.40 \\
SalUn &  0.89 $\pm$ 0.01 & 4.70 $\pm$ 0.78 & 92.80 $\pm$ 0.14 & 99.25 $\pm$ 0.19 & 97.85 $\pm$ 0.32 & 93.36 $\pm$ 0.41 \\
NGPlus &  0.91 $\pm$ 0.01 & 3.34 $\pm$ 0.53 & 92.71 $\pm$ 0.05 & 99.17 $\pm$ 0.13 & 96.49 $\pm$ 0.24 & 92.85 $\pm$ 0.52 \\
SCRUB &  0.91 $\pm$ 0.01 & 5.60 $\pm$ 0.74 & 92.90 $\pm$ 0.15 & 99.18 $\pm$ 0.17 & 98.5 $\pm$ 0.18 & 93.18 $\pm$ 0.63 \\


\cmidrule(lr){2-7}

SCRUB - SalUn &  0.90 $\pm$ 0.01 & 5.62 $\pm$ 0.74 & 92.92 $\pm$ 0.04 & 99.12 $\pm$ 0.18 & 98.53 $\pm$ 0.2 & 93.13 $\pm$ 0.66 \\
SCRUB - $\AND$ &  1.00 $\pm$ 0.00 & 4.30 $\pm$ 0.66 & 92.24 $\pm$ 0.19 & 98.83 $\pm$ 0.23 & 97.21 $\pm$ 0.31 & 92.57 $\pm$ 0.78 \\
\textbf{SCRUB - $\FOCUS$} &  0.90 $\pm$ 0.02 & 5.21 $\pm$ 0.82 & 92.66 $\pm$ 0.17 & 99.09 $\pm$ 0.18 & 98.12 $\pm$ 0.23 & 93.04 $\pm$ 0.65 \\

\cmidrule(lr){2-7}
SRL - $\AND$ & \textbf{0.69 $\pm$ 0.01} & 2.30 $\pm$ 0.47 & 92.18 $\pm$ 0.08 & 98.76 $\pm$ 0.21 & 95.45 $\pm$ 0.52 & 91.81 $\pm$ 0.41 \\
SRL - $\PROB$ & \textbf{0.76 $\pm$ 0.02} & \textbf{0.45 $\pm$ 0.36} & 91.69 $\pm$ 0.16 & 98.71 $\pm$ 0.22 & 93.61 $\pm$ 0.74 & 90.74 $\pm$ 0.47 \\
\textbf{SRL - $\FOCUS$} &  \textbf{0.77 $\pm$ 0.03} & \textbf{0.41 $\pm$ 0.50} & 91.71 $\pm$ 0.15 & 98.75 $\pm$ 0.23 & 93.57 $\pm$ 0.34 & 90.60 $\pm$ 0.14 \\

    \bottomrule
  \end{tabular}
  }
\end{table}

%% file: main/7_discussion.tex
Our experimental protocol considers 360 configurations: 5 add-on possibilities (none, salient, $\AND$, $\PROB$ and focus $\FOCUS$) over 3 unlearning methods, 2 model architectures, 2 datasets, and 2 types of forgetting scenarios with 3 ratios.
In general, our experimental results, not presented here due to a lack of space, confirm our observations over the architectures, the types of forgetting scenarios, and the datasets.
The ratio has an impact on the number of epochs.
As for the methods, NGPlus is dominated by SRL or SCRUB. NGPlus forces an augmentation of the gradient over the forget set $\sD_\FORGET$, which destroys the model capabilities along the epochs. A big loss on the forget set means that the model has to mostly equivocate on data that must be treated as such as if they belong to the test set.
This deludes basic membership inference attacks, showing they are not powerful enough. 
When the model systematically gives a wrong answer on some data, unlearning is detectable.
We recommend using NGPlus during one or two epochs only because the gradient ascent leads to unstable behavior of the resulting model.

The saliency masking of SalUn is justified in~\cite{Fan2023SalUnEM} by the argument of sparsity but does not take into account the variances while computing the mask.
Our paper gives two other justifications: feasible update and the noisy estimation of the gradients when processing by batch.
By combining these two arguments, we promote the idea that masking should prevent hazardous update directions.
Our focus vector is based on a simple statistical model that enable to derive the probability of a right parameter selection.
Finally, we show that the focus vector improves the performance of unlearning algorithms both in efficiency and efficacy.
It requires fewer steps to reach an acceptable rUA while conserving the accuracies and lowering the score of population-based MIA. 

The theoretical analysis assumes that the measured gradients on batches are a noisy representation of a true gradient with a simple Gaussian model.
Measured gradients certainly follow a more complex distribution.
However, the experiments already show the benefit of this simple assumption.
The last limitation of our work concerns the MIA metric.
There exist more powerful membership inference attacks specifically in unlearning, such as U-LiRa, but they require an unreasonable amount of computation power.
We chose to verify that the unlearned model lowers the performance of some basic MIAs compared to the initial model.
We expect that this also holds for U-LiRa knowing that basic attacks generally follow the same tendencies as U-LiRa as shown in~\cite{hayes2024inexact}.

%% file: main/appendix_proof.tex
\subsection*{Proof of Proposition ~\ref{prop:equilibrium}}

\begin{proof}
The first-order Karush-Kuhn-Tucker condition provides a constraint on a candidate solution $\bar{\theta}$: there exist $ \mu\geq0$ s. t. 
\[
\nabla U(\bar{\theta}) + \mu \cdot \nabla C(\bar{\theta}) = 0.
\]
Then, by multiplying component-wise by $\nabla C(\bar{\theta})$:
\[
\nabla U(\bar{\theta})\odot \nabla C(\bar{\theta}) + \mu \cdot\nabla C(\bar{\theta}) \odot \nabla C(\bar{\theta})= 0.
\]
We have
\(
\left(\nabla C(\bar{\theta}) \odot \nabla C(\bar{\theta})\right)_i = \left(\nicefrac{\partial}{\partial_{\theta_i}} C(\bar{\theta})\right)^2 \geq 0,
\)
so that
\[
\nabla U(\bar{\theta})\odot \nabla C(\bar{\theta}) = - \mu \cdot \nabla C(\bar{\theta}) \odot \nabla C(\bar{\theta}) \leq 0. \]
\end{proof}

\subsection*{Proof of Proposition~\ref{prop:guarantee}}

\begin{proof}
Knowing that $C(\theta_0)=0$ by definition~\eqref{eq:Constraint} and since $\langle\Delta, \nabla U(\theta_0)\rangle <0$ and $\langle\Delta, \nabla C(\theta_0)\rangle <0$ because the update is feasible, then, up to the first order in $\eta$, both functions $U$ and $C$ decreases along the direction $\Delta$. 
\end{proof}

\subsection*{Proof of Proposition~\ref{prop:bounding}}
\begin{proof}
The norm $\|m(\nabla U, \nabla C )\left(\theta_0\right)\|_0$ is the number of non-zero components, each of them being lower in absolute value to $\|\Delta(\theta_0)\|_{\infty}$:
\begin{align*}
    \|\theta - \theta_0 \|_q &= \|\eta \Delta(\theta_0) \|_q=\eta\left(\sum_i |\Delta(\theta_0)_i|^q \right)^{\nicefrac{1}{q}}\\
    & \leq \eta \|m(\nabla U, \nabla C )\left(\theta_0\right)\|_0^{\nicefrac{1}{q}} \|\Delta(\theta_0)\|_{\infty}.
\end{align*}
If $\sif(x,y) = \underset{z \in \{x, y\}}{\operatorname{argmin}} \,|z|$, one has both
\(\sif (x,y) \leq |x|\) and \(\sif (x,y) \leq |y|\) so that
\(
\|\Delta(\theta_0)\|_\infty \leq \|\sif(\nabla U, \nabla C )\left(\theta_0\right)\|_\infty \leq \|\nabla C (\theta_0)\|_\infty.
\)
\end{proof}

\subsection*{Proof of Proposition~\ref{prop:diragreeprob}}
\begin{proof}
    Knowing that the $i$-th component of the empirical gradient over a given batch equals $\hat{g}_{U,i}$, we have
\begin{equation*}
    \phi_{U,i}\doteq\Prob\left[g_{U,i} > 0 \,\big|\, \hat g_{U,i}\right]
    =\Prob\left[N_{U,i} < \hat g_{U,i} \,\big|\, \hat g_{U,i}\right]
    =\Phi\left(\frac{\hat g_{U,i}}{\sqrt{\Sigma_{U,(i,i)}}}\right).
\end{equation*}

The same holds for $\phi_{C,i}$. The estimation noises being supposed independent, $g_{U,i}$ and $g_{C,i}$ are both positive with probability $\phi_{U,i}\phi_{C,i}$.
Similarly, they are both negative with probability $(1-\phi_{U,i})(1-\phi_{C,i})$.
\end{proof}

\subsection*{Proof of Proposition~\ref{prop:Properties}}
\begin{proof}
    Case 1: For $p=\nicefrac{1}{2}$, we have
    \begin{align*}
        & \quad \phi_{U,i}\phi_{C,i}+(1-\phi_{U,i})(1-\phi_{C,i}) > \frac{1}{2} \\
        & \Leftrightarrow 2 \phi_{U,i} \left(\phi_{C,i} - \frac{1}{2}\right) > \phi_{C,i} - \frac{1}{2} \\
         & \Leftrightarrow \left\{\phi_{U,i} > \frac{1}{2} \cap \phi_{C,i} > \frac{1}{2} \right\}\quad \cup \quad \left\{\phi_{U,i} < \frac{1}{2} \cap \phi_{C,i} < \frac{1}{2} \right\} \\
        & \Leftrightarrow \left\{\hat g_{U,i} > 0 \cap \hat g_{C,i} > 0 )\right\}\quad\cup\quad\left\{\hat g_{U,i} < 0 \cap \hat g_{C,i} < 0 \right\} \\
        & \Leftrightarrow \hat g_{U,i} \hat g_{C,i} > 0.
\end{align*}

Case 2: The function $\Phi(x/\sigma)$ tends to an Heaviside step function when $\sigma\to0$:
\begin{equation*}
    \Phi(\nicefrac{x}{\sigma})\underset{\sigma\to0}{\to}
    \begin{cases}
    1 &\text{if}\, x>0\\
    0 &\text{if}\, x<0
    \end{cases}
\end{equation*}
Therefore
\begin{align*}
    \phi_{U,i}\phi_{C,i}&+(1-\phi_{U,i})(1-\phi_{C,i})\underset{\sigma\to0}{\to}\\
    &\begin{cases}
1\cdot 1 + 0\cdot 0 = 1, & \text{if } \hat g_{U,i} > 0 \text{ and } \hat g_{C,i} > 0\\
1\cdot 0 + 0\cdot 1 = 0, & \text{if } \hat g_{U,i} > 0 \text{ and } \hat g_{C,i} < 0\\
0\cdot 1 + 1\cdot 0 = 0, & \text{if } \hat g_{U,i} < 0 \text{ and } \hat g_{C,i} > 0\\
0\cdot 0 + 1\cdot 1 = 1, & \text{if } \hat g_{U,i} < 0 \text{ and } \hat g_{C,i} < 0.
\end{cases}
\end{align*}
\end{proof}

\subsection*{Proof of Proposition~\ref{prop:focusguarantee}}
\begin{proof}
    For any positive $\alpha$ and $\beta$, we write 
    \begin{align*}
        \mathbb{E}\left[\langle \Delta_{\FOCUS}, g_C \rangle\right] &=
        -\alpha \mathbb{E}\left[\langle f \odot \hat g_U, g_C \rangle\right]
        - \beta \mathbb{E}\left[\langle f \odot \hat g_C, g_C \rangle\right]\\
        &=-\alpha \mathbb{E}\left[\sum_i f_i \hat g_{U,i} g_{C,i}\right] -
        \beta \mathbb{E}\left[\sum_i f_i \hat g_{C,i} g_{C,i}\right].
    \end{align*}
    For the first term with $\alpha$, we condition on the observation $(\hat{g}_C,\hat{g}_U)$.
    From Prop.~\ref{prop:Properties} Case 1,  we know that $f_i > \frac{1}{2} \Leftrightarrow \hat g_{U,i} \hat g_{C,i} > 0$.
    We split the sum so that
    \begin{align*}
        \mathbb{E}&\left[\sum_i f_i \hat g_{U,i} g_{C,i} \,\bigg|\, \hat g_{U,i},\hat g_{C,i}\right] = 
        \sum_{i, f_i < \frac{1}{2}} f_i \hat g_{U,i} \hat g_{C,i} + 
        \sum_{i, f_i > \frac{1}{2}} f_i \hat g_{U,i} \hat g_{C,i}\\
        &= - \sum_{i, f_i < \frac{1}{2}} f_i |\hat g_{U,i} \hat g_{C,i}| + 
        \sum_{i, f_i > \frac{1}{2}} f_i \hat g_{U,i} \hat g_{C,i}\\
        &\geq - \frac{1}{2} \sum_{i, \hat g_{U,i} \hat g_{C,i} < 0}  |\hat g_{U,i} \hat g_{C,i}| + 
        \frac{1}{2} \sum_{i, \hat g_{U,i} \hat g_{C,i} > 0}  \hat g_{U,i} \hat g_{C,i}\\
        &= \frac{1}{2} \sum_{i} \hat g_{U,i} \hat g_{C,i} 
        =\frac{1}{2} \langle \hat g_U, \hat g_C \rangle
    \end{align*}
    Then, in expectation over $(\hat{g}_U,\hat{g}_C)$:
    \begin{equation*}
        \begin{split}
            \mathbb{E}\left[\sum_i f_i \hat g_{U,i} g_{C,i}\right] 
            &=\mathbb{E}\left[\mathbb{E}\left[\sum_i f_i \hat g_{U,i} g_{C,i} \,\bigg|\, \hat g_{U,i},\hat g_{C,i}\right]\right]
            \\
            &\geq \frac{1}{2} \mathbb{E}\left[\langle \hat g_U, \hat g_C \rangle\right] = 0.
        \end{split}
    \end{equation*}
    The second term with $\beta$ is trivially positive:
    \begin{equation*}
        \begin{split}
                    \mathbb{E}\left[\sum_i f_i \hat g_{C,i} g_{C,i}\right] &=  \mathbb{E}\left[\mathbb{E}\left[\sum_i f_i \hat g_{C,i} g_{C,i} \,\bigg|\,  \hat g_{C,i}\right]\right]  
                    \\
                    &= \sum_i \mathbb{E}\left[f_i \hat g_{C,i}^2\right] \geq 0. 
        \end{split}
    \end{equation*}
    This concludes that in expectation
    \[
    \mathbb{E}\left[\langle \Delta_{\FOCUS}, g_C \rangle\right] \leq 0.
    \]
    By the same token, interchanging the roles of $\alpha$ and $\beta$,
    we obtain $\mathbb{E}\left[\langle \Delta_{\FOCUS}, g_U \rangle\right] \leq 0$
    and the update $\Delta_{\FOCUS}$ is feasible in expectation.
\end{proof} 

\subsection*{Proof of Proposition~\ref{prop:PropertiesFocus}}
\begin{proof}

Case 1 is a corollary of Prop~\ref{prop:Properties}.
For Case 2, we have
\(
    \Phi(\nicefrac{x}{\sigma})\underset{\sigma\to\infty}{\to}\nicefrac{1}{2}
\),
therefore
\[\phi_{U,i}\phi_{C,i}+(1-\phi_{U,i})(1-\phi_{C,i})\underset{\sigma\to\infty}{\to} \nicefrac{1}{2}\cdot\nicefrac{1}{2} + (1-\nicefrac{1}{2})\cdot(1-\nicefrac{1}{2})=\nicefrac{1}{2}.\]
\end{proof}